\documentclass[11pt,twocolumn]{article}
\usepackage[dvips]{graphicx}
\usepackage{amssymb}
\usepackage [matrix, curve, arrow, frame, 2cell] {xy}
\usepackage[english]{babel}
\usepackage {latexsym}
\usepackage{xspace}
\usepackage{graphicx}
\usepackage{subfigure}
\CompileMatrices

\newcommand{\pdf}{\textit{pdf}\xspace}
\newcommand{\apriori}{\textit{a priori}\xspace}
\newcommand{\aposteriori}{\textit{a posteriori}\xspace}
\newcommand{\teruti}{\texttt{Ter Uti}\xspace}
\newcommand{\T}{\texttt{T}\xspace}
\newcommand{\N}{\texttt{N}\xspace}
\newcommand{\carottage}{\textsc{CarrotAge}\xspace}
\newcommand{\hmmd}{\textit{HMM2}\xspace}
\newcommand{\hmmu}{\textit{HMM1}\xspace}

\begin{document}

\title{Temporal and Spatial Data Mining with Second-Order Hidden
  Markov Models \thanks{Also in Soft Computing (2005) DOI 10.1007/s00500-005-0501-0}}
\author{Jean-Fran\c{c}ois Mari 
\thanks{LORIA - Inria Lorraine,  B.P. 239, 54506
 Vand{\oe}uvre-l{\`e}s-Nancy (France) email:
\texttt{jfmari@loria.fr}}
\and Florence Le Ber % \inst{1}  \inst{2}
\thanks{
CEVH, ENGEES, 1 quai  Koch, 67000
Strasbourg (France) email:
\texttt{fleber@engees.u-strasbg.fr}
}
}
\date{}
\maketitle
\begin{abstract}
In the frame of designing a knowledge discovery
system, we have developed stochastic models based on high-order hidden
Markov models. These models are capable to map sequences of data
into a Markov chain in which the transitions between the states depend
on the \texttt{n} previous states according to the order of the
model. We study the process of achieving information extraction from
spatial and temporal data by means of an unsupervised
classification. We use therefore a French national database related to
the land use of a region, named
\teruti, which describes the land use both in the spatial and
temporal domain. Land-use categories (wheat, corn, forest, ...) are
logged every 
year on each site regularly spaced in the region. They constitute a
temporal sequence of images in which we look for spatial and temporal
dependencies. 

The temporal segmentation of the data is done by means of a second-order
Hidden Markov Model (\hmmd) that appears to have very good capabilities
to locate stationary 
segments, as shown in our previous work in speech recognition. The
spatial classification is performed by defining a fractal scanning of
the images with the help of a Hilbert-Peano curve that introduces a
total order on the sites, preserving the relation of
neighborhood between the sites. We show that the \hmmd performs a
classification that is meaningful for the agronomists.

Spatial and temporal classification may be achieved simultaneously by
means of a 2 levels \hmmd that measures the \aposteriori probability to
map a temporal sequence of images onto a set of hidden  classes.
%\keywords{Temporal and spatial data mining, \teruti data, stochastic models, Markov chains.}
\end{abstract}

% \begin{keyword}
% % keywords here, in the form: keyword \sep keyword

% % PACS codes here, in the form: \PACS code \sep code
% Temporal and spatial data mining \sep \teruti data \sep 
% stochastic models \sep Markov chains.
% \end{keyword}

%\tableofcontents
%
%%%%%
\section{Context}
%%%%%

Mining sequential and spatial patterns is an active
area of research  in artificial intelligence. One basic problem in
analyzing a sequence of items is to find frequent episodes, i.e.
collections of events occurring frequently together. Early in  1995,
Agrawal~\cite{agrawal95,agrawal96} proposes non-numeric algorithms for
extracting regular 
patterns from temporal data. Conversely, we present in this
paper new numerical algorithms, based on high-order stochastic
models -- the second-order hidden Markov models (\hmmd) --  capable to
discover frequent 
sequences of events in temporal and spatial data. These algorithms
 were initially specified for speech recognition
purposes~\cite{mari85c} in our   laboratory. We show that, with minor
changes, they  
can extract spatial and temporal regularities that  can be explained by human
experts and may constitute some atoms of knowledge~\cite{brachman93}. 

The {\hmmd}'s are based on the probabilities  and statistics
theories. Their main 
advantage is the existence of   unsupervised training algorithms that
allow to estimate a model parameters from a  
corpus of observations and an initial model. Several criteria are
used~: the maximum likelihood criterium implemented in the well known EM 
algorithm\,~\cite{dempster77},  the maximum \aposteriori
criterium\,\cite{huo95}  and the maximum mutual information
criterium\,\cite{bahl86}.  Rabiner\,\cite{rabiner95} gives an extensive
description  of analytical learning methods for HMM. Other
algorithms\,\cite{quian90,robert93},     
based on stochastic likelihood maximization and Bayesian
estimation
have shown interesting results.  
The resulting model is
capable to segment each sequence in stationary and  transient
parts and to build up a classification of the data together with the
\aposteriori probability of this classification. This characteristic
makes the {\hmmd}'s 
appropriate to discover temporal and spatial regularities like it is
shown in various areas: speech recognition\,\cite{jelinek76,mari97b}
image restoration and segmentation\, \cite{benmiloud95},
genetics\,\cite{mury97,bize99,hergalant03a}, 
robotics\,\cite{aycard97a}, data mining\,\cite{berndt96,fine98,adibi01},
decision helping\,\cite{brehelin99}.  

We
focused our effort on two points: 1) the elaboration of a process of
mining  spatial and temporal dependencies aiming to the elicitation of
knowledge. This process always involves an unsupervised classification
of the data.   
2) The  specification of adequate visualization tools 
that give a synthetic view of the classification results to the
experts who have to interpret the classes and/or specify new experiments.

In this paper, we present our methodology and give some results in
the data mining of 
temporal and spatial data in the framework of the 
agricultural land use evolution.
We show on various examples
that  the {\hmmd}'s  are powerful tools for  temporal and spatial data
mining.  Another title of this paper could be: \emph{''How to
  understand what the land use talks to us''}. 

This paper is organized as follows. After an introduction (section 1),
Section 2 describes the
models that we used for classification purposes. Section 3
describes the agricultural data and  our attitude  in data mining
in collaboration with 
agronomists. The resulting  tied interaction yields to the production of a free
software named \carottage. The fourth section is the description of
two major applications of \carottage.  Section 5 is a conclusion.

\section{Modeling sequence dependencies with \hmmd}

\subsection{\hmmd definition}
\label{sec:def}
We define a second-order hidden Markov model by giving:
\begin{itemize}
\item {\(
{\bf S} = \left \{
s_1, s_2, \ldots, s_N, 
\right \} ,
\)} a finite set of  $N$ states ;
\item ${\bf A}$ a  matrix defining the transition probabilities 
      between the states:
\begin{description}
\item[
${\bf A} = \left ( a_{ij} \right )$] for a first order HMM (\hmmu),
\item[
${\bf A} = \left ( a_{ijk} \right )$]  for a second order HMM
(\hmmd) ;
\end{description}
      
\item {$\mathbf{b_i(.)}$} the distribution of observations associated to the
states $s_i$. This distribution may be parametric, non parametric or
even given by an HMM (see Figures~\ref{fig:super-model1} or
\ref{fig:super-model2}.
\end{itemize}

A Markov chain is defined over a set of states -- the crops in a field, or more generally the land-use categories in a place -- that are unambiguously
observed. The Markov  
chain specifies only one stochastic process, whereas in a HMM, the
observation of a land-use category is not uniquely associated to a
state $s_i$ but is rather a random variable that has a conditional
density $b_i(.)$ that depends on the actual state $s_i$~\cite{baker74a}. 
There is  a doubly stochastic process:
 
-- the former is hidden from the observer and is defined on a set
   of states;

-- the latter is visible. It produces an observation, the land-use
   of a parcel, at each time slot depending on the probability density 
   function (\pdf) that is defined on the  state in which the Markov
   chain stays at time $t$. It is often said that the Markov chain
   governs the latter. 

In this framework, we consider that the distribution of the country's
 land use 
 is  a Markov chain. The crop pattern at 
year $t$ depends at least  upon the crop pattern  the year before $(t-1)$
 or 2 years  before $(t -2)$.
 
\subsection{Automatic estimation  of a  \hmmd}

The estimation of an \hmmu  is usually done by the Forward Backward algorithm
which is related to the EM algorithm\,\cite{dempster77}. We have shown in
\cite{mari97a} that an \hmmd can  be
estimated following the same way. The estimation is an iterative
process starting with an initial model and a corpus of sequences of
observations that the \hmmd must fit even when the insertions, deletions and
substitutions of observations occur in the sequences.  
The very success of the HMM is based on their robustness:  even when the
considered data do not suit a given HMM, its use can give interesting results.
The initial model has
equiprobable transition   
probabilities and an uniform distribution in each state. At each step,
the Forward-backward algorithm determines a new model in which the
likelihood of the sequences of observation increases. 

Hence this
estimation process converges to a local maximum. Interested readers
may refer to \cite{dempster77,mari01a} to find more specific details
of the implementation of this algorithm.

If  $\N$ is the number of states and  $\T$ the sequence length, the
 Forward - Backward has a complexity of  
 $ \N^3 \times \T$ for an  \hmmd. 

The choice of the initial model has an influence on the final model
obtained by convergence. To assess this last model, we use the
Kullback-Leibler distance between the distributions associated to the
states~\cite{tou74}. Two states that are too close are merged and the
resulting model is re-trained. Agronomists do not interfere in the
process of designing a specific model, but they have a central
role in the interpretation of the results that the final model gives
on the data.

\subsection{Classification of a sequence using \hmmd}

The purpose of  pattern recognition is to specify as much models as
there are classes to recognize. As opposite to pattern recognition, we
do not have the 
knowledge of what to recognize but rather look for something regular
to extract, hence the name \emph{data mining}.
  
In the present work, we specify one HMM in order to model, in a more
simple way,  the  unknown behavior of a sequence. Each state captures a 
stationary behavior and represents a class where the observations are
drawn with a known \pdf. In an \hmmd the Forward-Backward
algorithm computes \aposteriori probability $a_{ijk}(t)$ 
that the Markov chain will be in state  $k$ at time $t+1$ knowing that
it has been 
in $j$ at  $t$ and in $i$ at  $t-1$. In  an \hmmu, the  Forward-Backward
algorithm computes \aposteriori probability $a_{ij}(t)$ on a smaller
period (2 years) that does not span most of land use successions
(ususally 2 or 3 years, sometimes 4).   
In the EM procedure, the \apriori transitions are calculated, at each
iteration, as being the mean of the \aposteriori transition
probabilities calculated with the current parameters.
 These \aposteriori probabilities can be
plot as a function of time and determines a fuzzy classification in
the states space.
This classification can be interpreted by a domain 
expert who can give it a meaning. Figure~\ref{celine} shows this
function and its interpretation by an agronomist.

Experiments results in speech recognition\,\cite{mari97a} show that
\hmmd's provide a better state occupancy modeling. In fact, the state
duration in an \hmmd is governed by two parameters, i.e, the
probability of entering a state only once, and the probability of
visiting a state at least twice, with the latter modeled as a geometric
decay. This distribution better fits a probability density of
durations than the classical exponential distribution of an \hmmu.

\subsection{Modeling the spatial data dependencies}

In the framework of image segmentation, Markov
random fields (MRF) offer probabilistic models in which a local  variable only
directly depends on  a few other neighboring variables. 
The estimation of a MRF and
the classification of an image  involve sophisticated and time consuming
algorithms~\cite{geman84}. 

More recently, iterative fuzzy clustering methods have been applied on
spatial data to find homogeneous regions.
Ambroise and Dang\,\cite{ambroise96}  proposes 
a variant of the EM algorithm -- called neighborhood EM (NEM) -- to
account for spatial proximity effects.  

\begin{figure}[htbp]

%% \begin{minipage}[b][4.cm][t]{.22\textwidth}
%% \includegraphics[width=\textwidth, height=3.cm]{./peano22.ps}
%% (a) 2x2~image
%% \end{minipage}
%% \begin{minipage}[b][4.cm][t]{.23\textwidth}
%% \includegraphics[width=\textwidth, height=3.cm]{./peano44.ps}
%% (b) 4x4~image~\\
%% \end{minipage}
%% ~\\
%% \begin{minipage}[b][4.cm][t]{.22\textwidth}
%% \includegraphics[width=\textwidth, height=3.cm]{./peano-image.ps}
%% ~\\(c) 6x6~image
%% \end{minipage}
%% \hfill
%% \begin{minipage}[b][4cm][t]{.23\textwidth}
%% ~\\
%%  \begin{xy}
%%   \entrymodifiers={++[o][F]}
%%   \xymatrix @C=1.5cm @R=1.5cm {  
%%   %% Ligne 1
%%   {1} \ar[r]   \ar@(ur, ul)[]  \ar[d] \ar[rd]
%%   & {2} \ar[l]  \ar@(ur, ul)[] \ar[d] \ar[ld] \\
%%   {3} \ar[r]   \ar@(dr, dl)[] \ar[u] \ar[ru]
%%   & {4} \ar[l]  \ar@(dr, dl)[] \ar[u] \ar[lu] \\
%%   }
%%   \end{xy}
%% ~\\~\\~\\
%% (d) Ergodic model
%% \end{minipage}
%% \caption{Definition of a total order in the plane by means of a
%%   Hilbert-Peano curve (a-c) and topology of the \hmmd used for
%%   segmentation (d). The hidden states are called: 1, 2, 3, 4.} \label{fig:ergo4}
\includegraphics{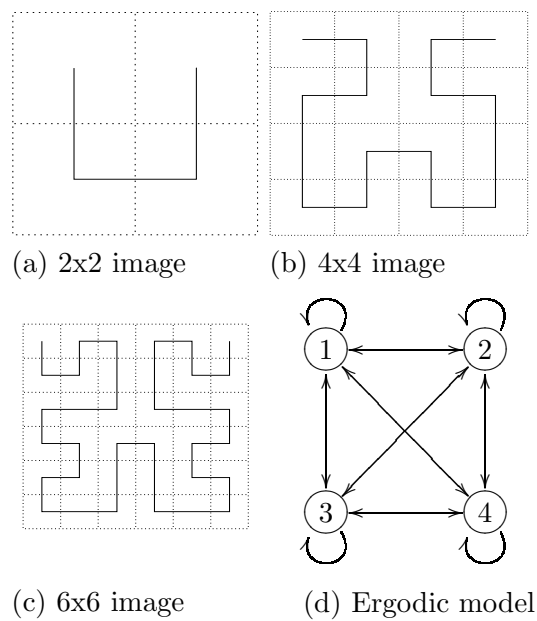}
\caption{Definition of a total order in the plane by means of a
  Hilbert-Peano curve (a-c) and topology of the \hmmd used for
  segmentation (d). The hidden states are called: 1, 2, 3, 4.} \label{fig:ergo4}
\end{figure}

Due to the low spatial resolution of our  data (see Section \ref{sec:teruti}), we
prefer to follow Benmilloud 
and Pieczynski~\cite{benmiloud95} and to use a rough clustering method
based on simple hidden Markov models. We have introduced a
total order in 
the image by means of a fractal curve -- a Hilbert-Peano curve -- that
scans the image 
 preserving the relation of
neighborhood between the sites. Two points that are close in the curve
are neighbor in the picture. But, the opposite is not true.
When a HMM specifies a segmentation of  the Hilbert-Peano curve,
it specifies also a region of the plane where the observations are
supposed stationary and come from the distribution associated to the
state.  The unsupervised classification is performed by an ergodic
model in which all the transitions between the states are possible
(see Figure~\ref{fig:ergo4}). 

\subsection{Modeling the spatio-temporal dependencies}

So far, the HMM classifies  the data on the basis of  their temporal
or spatial 
features. The
hidden partition is represented by the set of the HMM's states.

\begin{figure}[htbp]
%% \centering
%% \fbox{
%% \begin{xy}
%% \entrymodifiers={++[o][]}
%% \xymatrix @C=1cm {  
%% %% modele 1
%% \\
%%  {a} \ar[rr]   \ar@(ur, ul)[] & 
%% &{b} \ar[rr]  \ar@(ur, ul)[] & 
%% &{c}  \ar@(ur, ul)[]
%% \\
%% {1} \ar[r]   \ar@(ur, ul)[]  \ar[d] \ar[rd]
%% &{2} \ar[l]  \ar@(ur, ul)[] \ar[d] \ar[ld] 
%% &{1} \ar[r]   \ar@(ur, ul)[]  \ar[d] \ar[rd]
%% &{2} \ar[l]  \ar@(ur, ul)[] \ar[d] \ar[ld] 
%% &{1} \ar[r]   \ar@(ur, ul)[]  \ar[d] \ar[rd]
%% & {2} \ar[l]  \ar@(ur, ul)[] \ar[d] \ar[ld] 
%% \\
%%   {3} \ar[r]   \ar@(dr, dl)[] \ar[u] \ar[ru]
%% & {4} \ar[l]  \ar@(dr, dl)[] \ar[u] \ar[lu] 
%% & {3} \ar[r]   \ar@(dr, dl)[] \ar[u] \ar[ru]
%% & {4} \ar[l]  \ar@(dr, dl)[] \ar[u] \ar[lu]
%% & {3} \ar[r]   \ar@(dr, dl)[] \ar[u] \ar[ru]
%% & {4} \ar[l]  \ar@(dr, dl)[] \ar[u] \ar[lu]
%% \\
%% \\
%% }
%% \end{xy}
%% }
\includegraphics{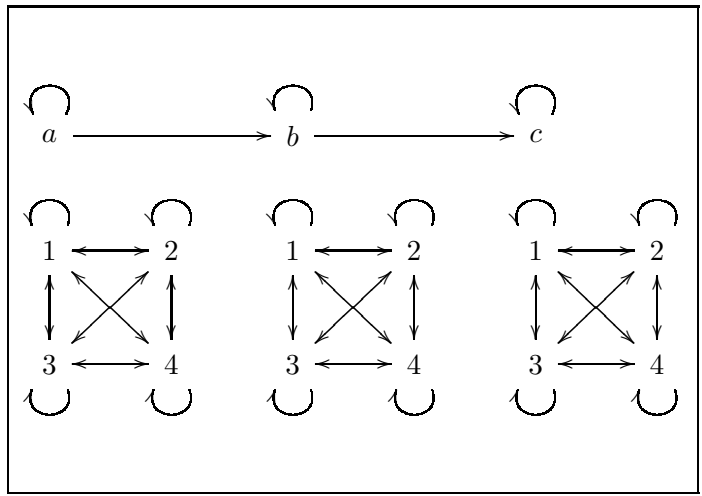}
\caption{Each state a, b, c of the temporal  master \hmmd is  a ergodic
  \hmmd depicted Figure~\ref{fig:ergo4} (d)}\label{fig:super-model1} 
\end{figure}

\begin{figure}[htbp]
 \centering
 %% \fbox{
%%  \begin{xy}
%%  \entrymodifiers={++[o][]}
%%  \xymatrix @C=1cm {  
%%  %% modele 1
%%  \\
%%    {1} \ar[r]   \ar@(ur, ul)[] 
%%  & {2} \ar[r]  \ar@(ur, ul)[]
%%  & {3}  \ar@(ur, ul)[]
%%  %% modele 3
%%  & {1} \ar[r]   \ar@(ur, ul)[] 
%%  & {2} \ar[r]  \ar@(ur, ul)[]
%%  & {3}  \ar@(ur, ul)[]    
%%  \\
%%  &
%%  &
%%    {a} \ar[r]   \ar@(ur, ul)[]  \ar[d] \ar[rd]
%%  & {b} \ar[l]  \ar@(ur, ul)[] \ar[d] \ar[ld] \\
%%  &
%%  &
%%    {c} \ar[r]   \ar@(dr, dl)[] \ar[u] \ar[ru]
%%  & {d} \ar[l]  \ar@(dr, dl)[] \ar[u] \ar[lu] \\
%%  %% modele 4
%%    {1} \ar[r]   \ar@(ur, ul)[] 
%%  & {2} \ar[r]  \ar@(ur, ul)[]
%%  & {3}  \ar@(ur, ul)[]
%%  %% modele 5
%%  & {1} \ar[r]   \ar@(ur, ul)[] 
%%  & {2} \ar[r]  \ar@(ur, ul)[]
%%  & {3}  \ar@(ur, ul)[]    
%%  }
%%  \end{xy}
%%  }
\includegraphics{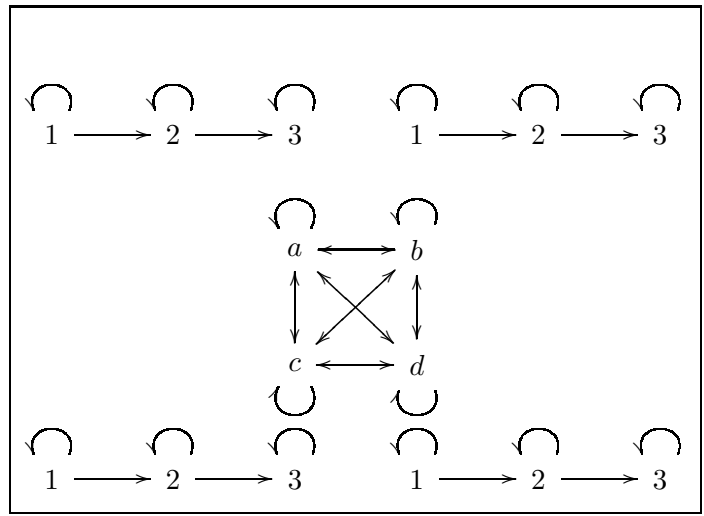}
 \caption{Each state a, b, c, d of the master \hmmd is a temporal \hmmd
   with states  1, 2, 3}\label{fig:super-model2}
 \end{figure}

We are now
interested in finding how both the temporal and spatial features of a
point in an image (a pixel) 
may interact to achieve a clustering in which the probability of the
hidden partition  depends in a coherent way both on the temporal and
spatial features. We still keep a \emph{Bayesian} point of view of the
classification. 
We have processed our spatio-temporal data following  two different ways: 
\begin{enumerate}
\item We have a sequence of  images, one per time slot. For example, 
 a site in an image is labeled by its  land-use category
(see Figure~\ref{fig:super-model1}).
\item We have one image. A site in the image is labeled by its temporal
  sequence of  
land-use categories (see Figure~\ref{fig:super-model2}).
\end{enumerate}

In both cases,  the probability of the
observation of a particular sequence  is given by an
HMM\,\cite{rabiner95,mari01a}.

In both experiments, the \hmmd that we use for image segmentation is an
ergodic model (see Figure~\ref{fig:ergo4}-d).

The data are structured as a matrix in which the rows represent the
sites ordered by the Hilbert-Peano curve and the columns represent the
time slots. In case (1), the sequence of images is represented by the
sequence of columns. The matrix is then processed columns  by columns
whereas in case (2), the matrix is processed rows by rows.
We define a master \hmmd whose states are in fact classical \hmmd. We
call them \emph{super states}. The
master \hmmd 
``generates'' observations that are vectors whose probability is given
by the smaller \hmmd associated to the super states. In the
Figures ~\ref{fig:super-model1} and ~\ref{fig:super-model2}, the super
states of the 
master \hmmd are called \emph{a,b,c,...}  whereas the states of the
\hmmd associated to the super states are called \emph{1, 2, 3,
  ...}. In Fig.~\ref{fig:super-model1}, the master HMM is temporal. In
Fig.~\ref{fig:super-model1}, it is spatial.

Our master  \hmmd can be compared to those coming from  the work of
 Saon~\cite{saon96,saon97b} that uses \hmmu's  
 for the
recognition of unconstrained handwritten words. In Saon's work, a master
\hmmu generates vectors representing columns of pixels. The sequence of
columns represents the matrix in which the handwritten word is
drawn. In our model whose topology is given
 Figure~\ref{fig:super-model2}, the  observation vector does not 
 represent a spatial direction 
but is rather a time dimension. Similar models have been used in
various applications. Fine~\cite{fine98} has defined in 1998
 Hierarchical HMM for 
 learning multi-resolution structures of natural English text.
 Adibi~\cite{adibi01} uses Similar Layered HMM 
 for analyzing and predicting the flow of packets in a
 network. 

In our experiments, we have considered the case of one image in which
the sites are labeled by a temporal sequence of land-use categories (case
2). The complexity of the Forward-backward algorithm can be assessed as
follows. Assume that:
\begin{itemize}
\item $N_M$ is the number of states in the master
\hmmd ;
\item  $N_s$, the number of states in the conventional \hmmd
associated in the \emph{super states} ;
\item  $R$ is the number of sites ;
\item 
$T$ the number of time slots.
\end{itemize}
The Forward-backward algorithm has a complexity of $N_M^3 R$ and involves
the computation of $N_M R$ probabilities given by the conventional
{\hmmd}'s whose complexity  is $N_s^3 T$. Then, the overall computation
stays polynomial. A very similar  demonstration stands for a
sequence of images (case 1).

\section{HMM for mining spatio-temporal data}

\subsection{Introduction}

One of the aspects of the data mining is to give a representation of
the data that an expert can interpret. Classification is the most
popular way to have a synthetic view of how the data are
structured. Our purpose is to build a partition -- called the hidden 
partition -- in which the inherent noise of the data is withdrawn as
much as possible. In the mining of spatio-temporal data, we are
interested in extracting homogeneous classes both in temporal and spatial
domains, and having a clear view of how are the transitions between
these classes. 

The process of data
mining does not stop as soon as some regularities have been extracted
but goes through several model specifications that incorporate the
units of knowledge that were previously extracted from the data.
In this process, the domain expert -- i.e. the agronomist
-- plays a central role. Its task is to interpret the results of each
   training of various models, and to suggest new models.

Our interaction with the agronomist went through 3 steps:  

\begin{enumerate}
\item
we first have
proposed to segment the data in the temporal domain with the help of a
simple 
\hmmd (see Figure~\ref{fig:topo6}). A tied interaction leads to the
definition of more complicated models as in Figure~\ref{fig:topo3} in
which the agronomists can extract and quantify information on
successions of land-use categories.

\item 
We next have proposed to segment the whole country on a spatial
basis and get the 2-D Map (Figure~\ref{fig:lorraine}). This map helps
the agronomist to distinguish  the actual regions on the basis of
actual successions of land-use categories.

\item
Finally, we have proposed the spatio-temporal segmentation to unify the
two former methods. And, we have used it to help the interpretation of
satellite images (see Section~\ref{sec:satellite}).
\end{enumerate}

\subsection{An example of spatio-temporal data: the \teruti data}
\label{sec:teruti}

\begin{figure*}[htbp]
    \centering
\subfigure[the basic grid in 3820 meshes]
{
  \includegraphics[height=6cm]
  {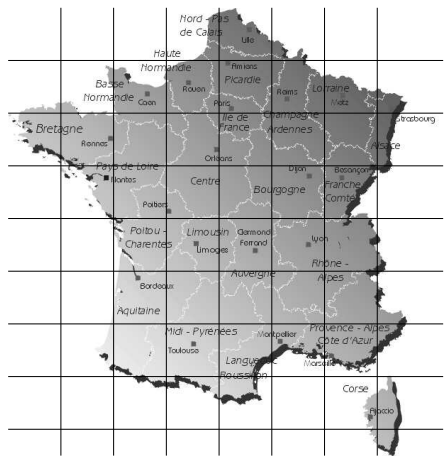} 
} 
\subfigure[the 4 air photos in a mesh]%[the relative position of the 4 air photographies in a mesh]
{
\includegraphics[height=6cm]
{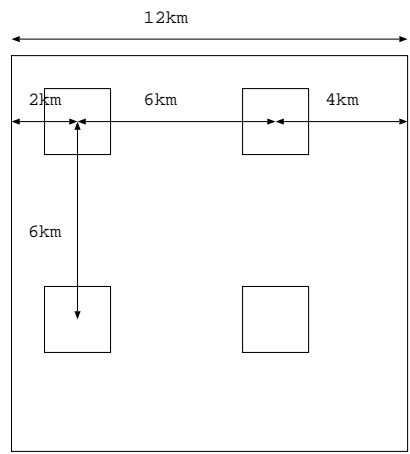}
} 
\subfigure[the air photography and its 6x6 grid]
{
  \includegraphics[height=6.6cm]  
  {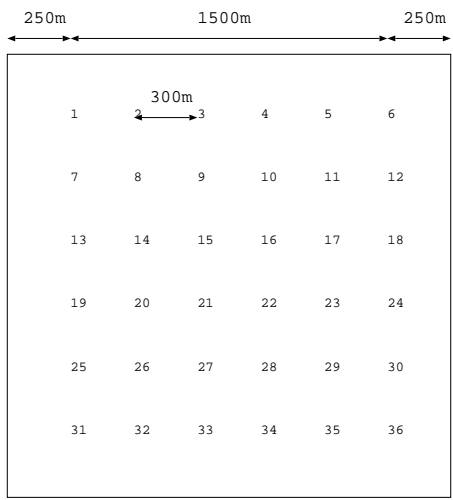}
  \includegraphics[height=6cm]  
  {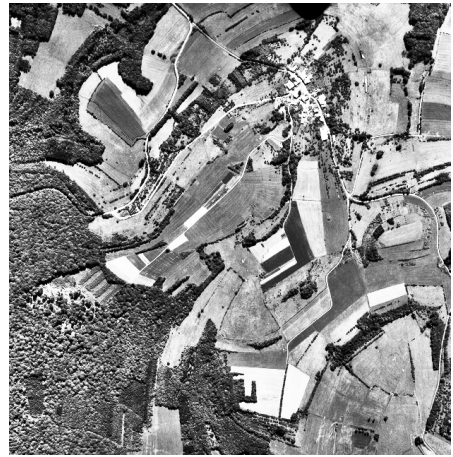}
}
    \caption{Description of the \teruti data: 3820 meshes square
      France, 4 air photographies are sampled in a mesh, a 6x6 grid
      determines 36 sites.} 
    \label{fig:teruti}
  \end{figure*}

The \teruti data are collected by the French agriculture
administration on the whole metropolitan territory. They represent the
land use of the country on a one year basis.  A first sample, done by
the IGN\footnote{Institut G\'eographique National.} consists to
select aerial photographies to cover a part of the entire country
(see  Figure~\ref{fig:teruti}). Two levels of resolution are
achieved. Each of the 3820 meshes contains 4 air photographies and 
a air photography 
covers only a square of 2 km. On each
photography, a 6 by 6 grid determines 36 sites that are inquired every
year in June. The fractal curve that we use for the scanning
takes into account these two levels of spatial resolution. A
Hilbert-Peano curve orders the \verb|32x32| photographies that cover
the region under study -- \emph{ie} Lorraine or Midi-Pyr{\'e}n{\'e}es
-- whereas an adapted curve (see Figure~\ref{fig:ergo4}-c) scans the
\verb|6x6| grid of sites in the photography. 
The land-use category of these sites (wheat, corn,
forest, ...)  is logged in a matrix in which the rows are the sites of
the country (30000 for the Lorraine) and the columns the time slots
(from 1992 to 2000). In our study, there are 49 modalities for
land-use categories~\cite{benoit01a}. One \teruti site represents
roughly 100 hectares. 

%\section{Temporal classification} \label{sec:temporal}
\subsection{Models for the temporal classification}
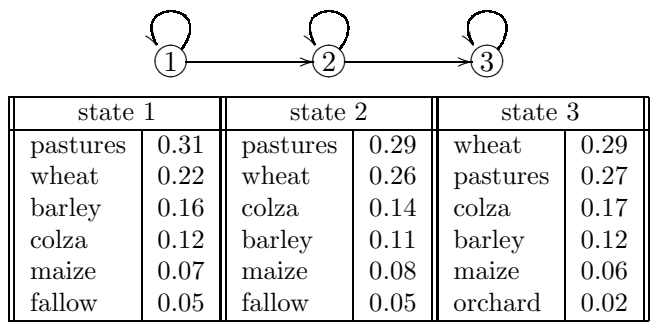
\begin{figure}[htp]
\vspace{0.2cm}
\centering
\mbox{
\begin{xy}
\entrymodifiers={+[o][F]}
\xymatrix @C=1.7cm {
%% Ligne 1
  {1} \ar[r]   \ar@(ur, ul)[]
& {2} \ar[r]  \ar@(ur, ul)[]
& {3}  \ar@(ur, ul)[]
}
\end{xy}
}

\vspace{0.2cm}
\small{
\mbox{
\begin{tabular}{||l|l||l|l||l|l||} \hline
\multicolumn{2}{||c||}{state 1} 
& \multicolumn{2}{|c||}{state 2} 
& \multicolumn{2}{|c||}{state 3} \\ \hline
pastures  & 0.31  & pastures  & 0.29  &  wheat & 0.29 \\
wheat & 0.22  &  wheat & 0.26 &  pastures & 0.27 \\
barley & 0.16 &  colza & 0.14 &  colza & 0.17 \\
colza & 0.12  &  barley & 0.11 &  barley & 0.12  \\
maize & 0.07  &  maize & 0.08 &    maize & 0.06 \\ 
fallow & 0.05 &  fallow & 0.05  & orchard & 0.02 \\ \hline
\end{tabular}
}
}
\caption{Model 1:  \hmmd performing a segmentation in 3 periods in which the observations are supposed stationary. We notice the 
the progression of the  wheat and the disappearance of the fallow
lands. The hidden states are called  2, 3 and 4. }
\label{fig:topo6}
\end{figure}

In a first study, we have been interested in the extraction of
temporal segments in which the distribution of the land-use categories
is
stationary. To do so, we have specified a \hmmd with 2 or 3 states with a
left to right, self loops topology (see Figure~\ref{fig:topo6}). This
means that we attempt to 
capture 2 or 3 periods of evolution in the land use dynamics.

%\vspace{1 cm}

We are interested in finding 3 stationary agricultural periods. 
Figure~\ref{fig:topo6} illustrates the results of three temporal
clusterings.

Between 1992 and 1999, the country went through three different states
with different distributions. The agronomists recognize the 
progression of wheat and the disappearance of the fallow land, two
processes that are related to the EC agricultural policy.

\begin{figure} [htbp]
\begin{center}
\mbox{ 
\begin{xy}
\entrymodifiers={++[o][F]} 
\xymatrix @C=1cm {  
%% Ligne 1
{1} \twocelltail{\dir{<}} \ar[r]  \ar@{-}[d] \ar@(ur, ul)[] \ar@{-}@/_0.7cm/[dd] 
\ar[rd] \ar[ddr]
&{2} \ar[r]  \ar@{-}[d] \ar@(ur, ul)[]\ar@{-}@/_0.7cm/[dd] 
\ar[rd] \ar[ddr]
& {3}  \ar@(ur, ul)[]  \ar@{-}[d] \ar@{-}@/_0.7cm/[dd] 
\\ %% ligne 2 = bl{\'e}
{wheat_1} \ar[r]  \ar@(ur, ul)[] \ar[ru] \ar@{-}[d] \ar[rd]
&  {wheat_2}\ar[r]  \ar@(ur, ul)[] \ar[ru] \ar@{-}[d] \ar[rd]
& {wheat_3} \ar@(ur, ul)[] \ar@{-}[d]
\\ %% ligne 3 = mais
{maize_1} \ar[r]  \ar@(ur, ul)[] \ar[ru] \ar[uur] 
& {maize_2} \ar[r]  \ar@(ur, ul)[] \ar[ru] \ar[uur]
& {maize_3}  \ar@(ur, ul)[]  
 %% etc ...
%% ...  & ...   & ...
%*{\vdots} & *{\vdots}  & *{\vdots} 
}
\end{xy}
}
 \caption{Model 2: the states denoted  1, 2 and 3 are associated to a
 distribution of land-use categories, as opposite to the states denoted with a
 specific land-use category. The number of columns determine the number of time
 intervals (periods). A connection without arrow means a bi directional 
 connection.}
\label{fig:topo3}
\end{center}
\end{figure}
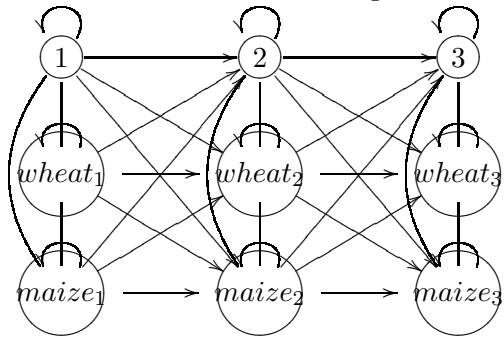

An \hmmd cannot  measure the probability of a
 succession of three land-use  categories because the states between which the
 transition probabilities stand  represent a distribution. To
 tackle this problem, we have defined a specific state, called the \emph{Dirac
 state}, whose distribution is zero except on a particular
 land-use category. Therefore, the transition probabilities between the  Dirac
 states measure the probabilities between the land-use categories
 during a three years period. 
Figure~\ref{fig:topo3} shows the topology of a \hmmd that has two
 kinds of states: Dirac states associated to the most frequent
 land-use categories (wheat, maize, barley,\,\ldots) and container states
 associated to a distribution of land-use categories like it is usually done in
 HMM modeling framework.

\subsection{Spatial classification}

\begin{figure}[h]
\subfigure[\hmmd classification] 
{
\includegraphics[height=6cm]
{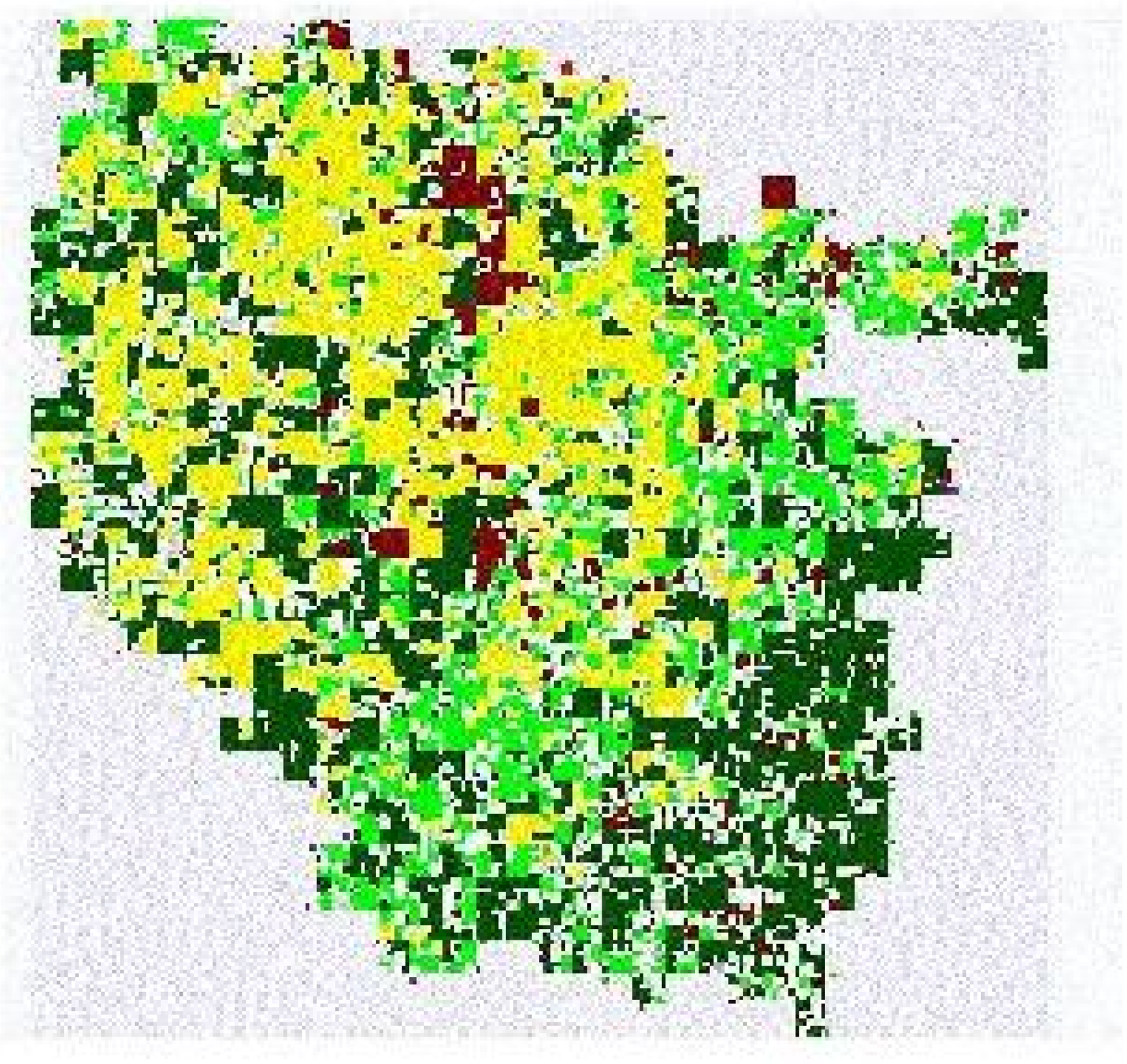}\label{figure11}
}
\subfigure[Map based on satellite images] 
{
\includegraphics[height=6cm]
{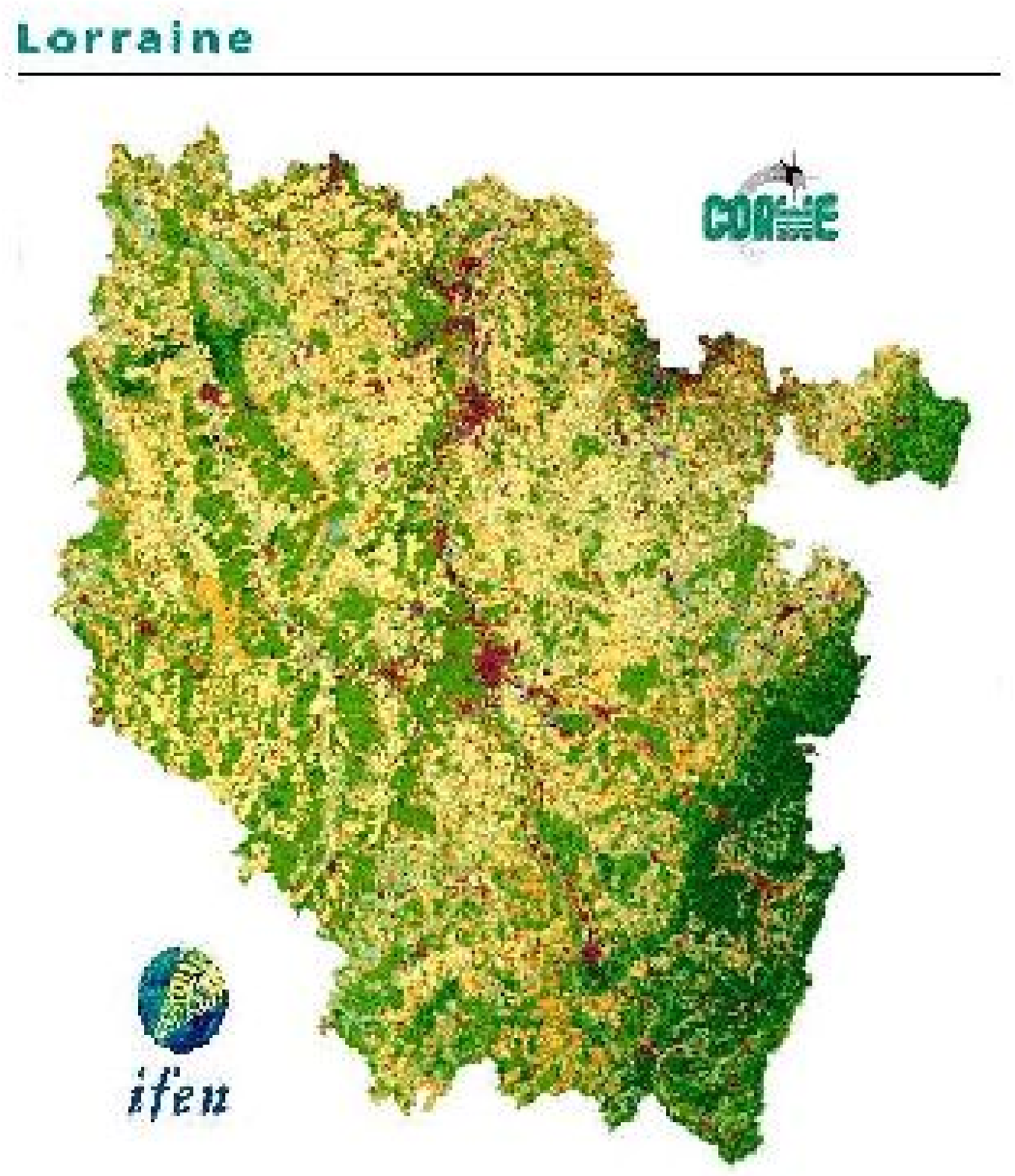} \label{figure12}
}
\caption{\hmmd classification of Lorraine \teruti data and  comparison
  with a map of the same region based on satellite images (IFEN,1993).
}\label{fig:lorraine}
\end{figure}

We did the first spatial experiment on Lorraine  data: 7 images of
30000 pixels, a 
pixel has 49 modalities corresponding to its land-use category at a time slot
(between 1992 and 1998). We trained an ergodic \hmmd (see
Figure~\ref{fig:ergo4}) with 5 
states on a corpus of these 7 images. After 10 iterations the
resulting model exhibits 5 distributions of land-uses:
\begin{itemize}
\item a state with a majority of houses, rivers, forests that follow
the valleys;
\item a state with a majority of forest (98 \%) located in Vosges and
Meuse;
\item a state with a majority of pastures (30 \%), forest (20
\%) and fodder (6 \%) located in the breeding countries; 
\item a state with a majority of cereals (wheat, maize,
barley,\,\ldots);
\item a state with a majority of pastures, orchards in the bottom of
valleys and mountains.
\end{itemize}
Picture~\ref{figure12} compares the segmentation performed by a
\hmmd and a satellite image in which the resolution is four times higher.

\section{Applications: \carottage in use}
%\subsection{Introduction}
\carottage\footnote{http://www.loria.fr/\~~jfmari/App/} is a free
software under a GPL licence\footnote{Gnu Public Licence}
that takes as input an array
of discrete data -- the rows represent the spatial sites and the columns
the time slots --  and builds a partition together with its \aposteriori
probability. This probability may be plot as a function of time and is
a meaningful feature for the expert looking for stationary and
transient behaviors of the data. \carottage is written in C++ and runs
under  Unix
and X11R6 systems. It is now used by agronomists and
geneticians~\cite{hergalant03a} 
without any assistance of the designers.
 
\subsection{Crop rotations in the Seine river watershed}

For thirty or forty years, the hydro-system of the Seine river has been
gradually degraded, regarding water quality and biological population,
due to the human activities (domestic, industrial, agricultural)
\cite{meybeck98}. The nitrate contamination of 
cave and surface 
waters is mainly caused by the evolution of agricultural activities,
and related to their 
nature and to their organization inside the river watershed. The aim
of the interdisciplinary research program PIREN-Seine (Programme
Interdisciplinaire de Re\-cher\-che  en ENvironnement sur la
Seine)~\cite{mignolet03a} is thus to develop a tool for  
forecasting water quality in the Seine river watershed, based on
assumptions upon agricultural changes. In 
this research framework, members of the IN\-RA team in Mirecourt
analyze the agricultural activities in the watershed, their dynamics
and their spatial organizations. They particularly
focus on the crop (temporal) rotations that are able to explain the
risk of nitrate loss.  

We present an example of the 
approach and the results obtained on a small agricultural region from
the north-east of 
France, the PRA (Petite R\'egion Agricole) St-Quentinois et
Laonnois. 

\begin{figure*}[htb]
\begin{center}
\includegraphics[width=\textwidth]{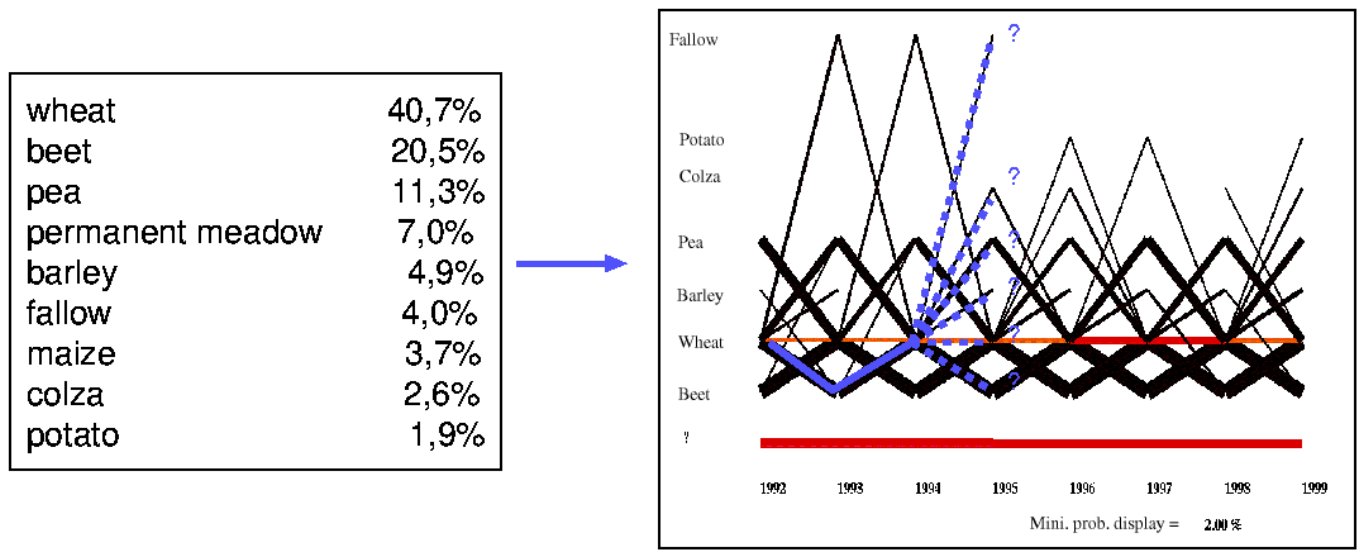}
 \caption{Results obtained on the \teruti data of PRA St-Quentinois et Laonnois
 (Aisne, years 1992-2000).}\label{celine}
\end{center}
\end{figure*}  

Several models have been used  in order to:
\begin{itemize}
\item compute the mean crop distribution in a given period, here from
1992 to 1999 (Figure \ref{celine}, left) ; the states of the \hmmd are
only container states. ;
\item view the main annual transitions between crops (Figure
\ref{celine}, right) ; the states of the \hmmd are container and Dirac
states (beet, wheat, pea, etc.).
\end{itemize}

 These results are analyzed as follows. One can see that the main
rotation heads -- beet and pea -- are generally followed and preceded
by wheat. It is thus assumed that the rotations contain triples from
the types \emph{wheat-beet-wheat} or \emph{wheat-pea-wheat}, and that
they are 4-years rotations, either \emph{?-wheat-beet-wheat} or
\emph{?-wheat-pea-wheat}. But these results do not make possible to
determine the crop (denoted with ?), which is before or after these
triples of crops. In Figure \ref{celine} (left) the dashed lines
represent the possible transitions between the triple
\emph{wheat-beet-wheat} and the other crops: beet, pea, wheat, barley,
colza or fallow.

Other models have been used for searching all types of crop rotations
in this small region. The same analysis has been done for all small
agricultural regions in the Seine watershed. The regions are then
clustered according to their main crop rotations and their
evolutions.
% This classification is a meaningful
%result for specifying  simulation models  of nitrate loss.
These results are meaningful for specifying  simulation models of nitrate loss and thus forecasting water quality in the Seine watershed.

\subsection{Helping satellite images
  interpretation in the Midi-Pyr{\'e}n{\'e}es region} \label{sec:satellite}

The first spatio temporal  experiment has been done on \teruti data
coming from the 
Midi-Pyr{\'e}n{\'e}es region. 

Our approach has been used by researchers of the INRA research center
in Toulouse  
working on the forecast of irrigation needs in the Midi-Pyr{\'e}n{\'e}es 
region (south-west of France). Usually, irrigation needs are estimated
thanks to annual land-use maps based 
on satellite data~\cite{casterad98}. This method is not satisfying
since all data are not available at 
the moment when the forecast has to be done: the satellite images
obtained in spring 
do not allow to distinguish all the crops of a given region. But if
the crop rotations are known, it can help to distinguish the crops,
based on the land-use map of the year before. Knowing the crop in a
plot at year $n-1$, the possible crops in the same plot 
at year $n$ can be infered, and their number reduced thanks to the
satellite images of spring.

\begin{figure}[hbt]
\begin{minipage}[t][7cm][t]{.40\textwidth}
\includegraphics[width=0.9\textwidth] {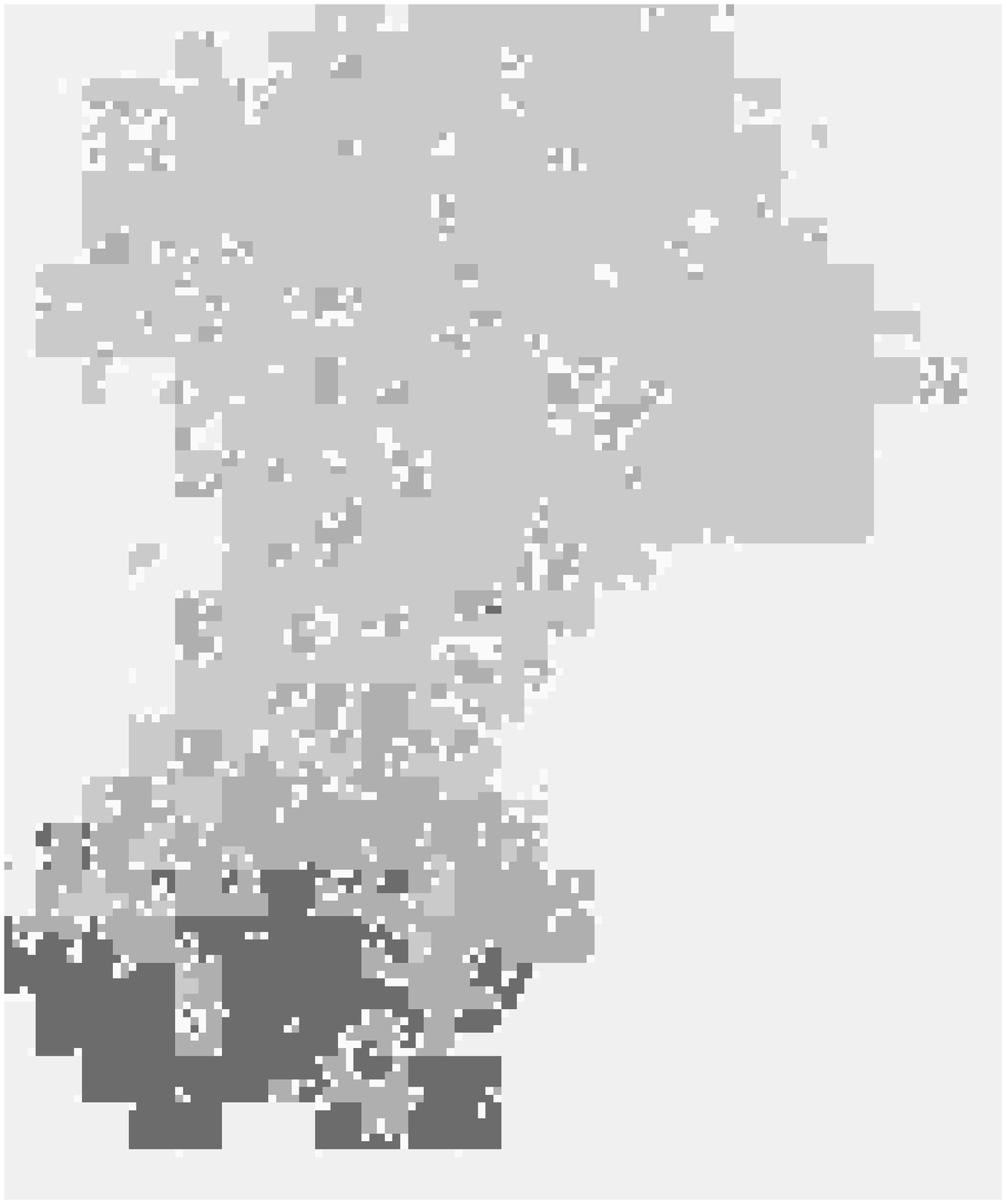}
\label{fig:midi-pyr}
\end{minipage}
\begin{minipage}[b][7cm][t]{.42\textwidth}
\small{
\begin{tabular}{||l|l||} \hline 
\multicolumn{2}{|c|}{around 1992} \\ \hline
0.099   & wood + wood + wood \\
0.089   & maize + maize + maize \\
0.082    & pastures + pastures + pastures \\
0.034    & wheat + sunflower + wheat \\
0.028    & vines + vines + vines \\
0.026    & sunflower + wheat + sunflower \\
0.018    & fallow + fallow + fallow \\ \hline
\end{tabular} 
\\
\begin{tabular}{||l|l||} \hline 
\multicolumn{2}{|c|}{around 2000} \\ \hline
0.105    & maize + maize + maize \\
0.099    & wood + wood + wood \\
0.081    & pastures + pastures + pastures \\
0.056    & wheat + sunflower + wheat \\
0.037    & sunflower + wheat + sunflower \\
0.027    & vines + vines + vines \\
0.023    & fallow + fallow + fallow \\ \hline
\end{tabular}
}
%\caption{R{\'e}partition des cultures} \label{tab1-gers-pyr}
\end{minipage}
\caption{Map of the crop rotations in the Midi-Pyr{\'e}n{\'e}es region and
evolution table in the north area (years 1992-2000).}\label{fig:pyr_gers}
\end{figure}

   In this framework, our approach has been used for a spatio-temporal
segmentation of the 
\teruti data of the Midi-Pyr{\'e}n{\'e}es region. We have specified
two different \hmmd. The master 
one is an ergodic \hmmd that has to segment an image like in the former
experiment. In each of the 5 states, we have defined a temporal 3
states \hmmd like in Figure~\ref{fig:topo3}. The territory is separated
into three homogeneous areas according to the crop rotations
(Figure~\ref{fig:pyr_gers}): the south area (black) is mainly 
mountains (Pyr{\'e}n{\'e}es), the middle area (dark grey) is covered with
alpine pastures and 
forests, and the north area (grey) is cultivated. The study then
focused on the north 
area, for analyzing the evolution of the crop rotations. The table of
Figure~\ref{fig:pyr_gers}  shows the 
growing of the wheat-sunflower rotation and of the maize
mono-culture. Eventually, the 
transition probabilities between crops (Dirac states) have been
computed for the ten 
years where \teruti data were available. These transition
probabilities have then been used with a land-use map of the year
$n-1$ and an April satellite image in the year $n$, in order to
estimate the land-use map of the year. This estimation is then
evaluated 
with a land-use map of the year built \aposteriori~\cite{mesmin02}.
                                  
   In this example, the models defined for spatial and temporal
segmentation have been used in a complementary way. The first ones
allow to define homogeneous stable  areas regarding the crop
rotations, while the last ones allows a more specific study of 
each area. 

\section{Conclusion}

We have described a clustering method on spatial and temporal data
based on second-order Hidden Markov Models.  The \hmmd maps the
observations into a set of 
states generated by a second order Markov chain. The classification is
performed, both in time domain and spatial domain, by using the
\aposteriori probability that the stochastic process is in a
particular state assuming a sequence of observations. We have shown
that spatial data may be re-ordered using a fractal curve that
preserves the neighboring information. We adopt a Bayesian point of
view and measure the temporal and the spatial
variability with the \aposteriori probability of the mapping. Doing
so, we have a coherent processing both in temporal and spatial
domain.  This approach appears to be valuable for spatio-temporal
data mining. Indeed,   the domain experts  specify  the models and
the \hmmd performs an
unsupervised   clustering process. Then the domain experts interpret
the classification and find in the 
results an objective information. Further works will include a
comparison with Markov Random  Fields.  The improvements of \carottage
will be driven by the applications needs (Agronomy, Genetics).

\section*{Acknowledgment}
This paper has been published in Soft Computing (2005). The original
publication is available at www.springerlink.com under 
DOI \texttt{10.1007/s00500-005-0501-0}.

\bibliography{/users/orpailleur/jfmari/Papers/Cassini2002/supcassini,/users/orpailleur/jfmari/Biblio/biblioall,/users/orpailleur/jfmari/Biblio/bibliomari,/users/orpailleur/leber/Biblio/agro}

\end{document}